\documentclass[letterpaper]{article} 
\usepackage{aaai24}  
\usepackage{times}  
\usepackage{helvet}  
\usepackage{courier}  
\usepackage[hyphens]{url}  
\usepackage{graphicx} 
\usepackage{natbib}  
\usepackage{caption} 
\usepackage{algorithm}
\usepackage{algpseudocode}
\usepackage{amsmath}
\usepackage{amssymb}
\usepackage{array}
\usepackage{bm}
\usepackage{breqn}
\usepackage[subrefformat=parens]{subcaption}
\usepackage{comment}

\usepackage{newfloat}
\usepackage{listings}
\DeclareCaptionStyle{ruled}{labelfont=normalfont,labelsep=colon,strut=off} 
\floatstyle{ruled}
\newfloat{listing}{tb}{lst}{}
\floatname{listing}{Listing}

\title{BSED: Baseline Shapley-Based Explainable Detector}
\author {
    Michihiro Kuroki\textsuperscript{\rm 1},
    Toshihiko Yamasaki\textsuperscript{\rm 1}
}
\affiliations {
    \textsuperscript{\rm 1}The University of Tokyo\\
    kuroki@cvm.t.u-tokyo.ac.jp, yamasaki@cvm.t.u-tokyo.ac.jp
}

\begin{document}

\maketitle

\begin{abstract}
Explainable artificial intelligence (XAI) has witnessed significant advances in the field of object recognition, with saliency maps being used to highlight image features relevant to the predictions of learned models. Although these advances have made AI-based technology more interpretable to humans, several issues have come to light, as some approaches present explanations irrelevant to predictions, and cannot guarantee the validity of XAI (axioms). In this study, we propose the Baseline Shapley-based Explainable Detector (BSED), which extends the Shapley value to object detection, thereby enhancing the validity of interpretation. The Shapley value can attribute the prediction of a learned model to a baseline feature while satisfying the explainability axioms. The processing cost for the BSED is within the reasonable range, while the original Shapley value is prohibitively computationally expensive. Furthermore, BSED is a generalizable method that can be applied to various detectors in a model-agnostic manner, and interpret various detection targets without fine-grained parameter tuning. These strengths can enable the practical applicability of XAI. We present quantitative and qualitative comparisons with existing methods to demonstrate the superior performance of our method in terms of explanation validity. Moreover, we present some applications, such as correcting detection based on explanations from our method.
\end{abstract}

\section{Introduction}
AI-based object recognition plays an important role across various domains. In the medical field, for instance, AI-based object recognition systems aid physicians in the enhanced and precise diagnosis of diseases. Nonetheless, the black-box nature of AI poses challenges in confidently utilizing such systems. Consequently, the interpretability of AI has drawn prominent academic attention, with extensive research~\cite{DARPA, XAISurvey2} conducted on the topic. This process is critical for the widespread social acceptance of AI systems.\\
\indent Extensive research has been conducted on explainable artificial intelligence (XAI) for image classification tasks. Typically, pixel-wise feature attributions are calculated based on classification confidence scores and depicted as a saliency map. The feature attribution can be interpreted as an importance score of each pixel for classification. Several different approaches can be employed to calculate the feature attribution. Back-propagation-based methods utilize gradients of the neural network in a learned model, while activation-map-based methods use feature maps in the convolutional neural network layer. Because these methods are dependent upon network architecture, prior knowledge regarding the model is required. In contrast, perturbation-based methods take samples of perturbated input images and their corresponding output scores to calculate the feature attributions. These methods are independent of network architecture, and can be applied in a model-agnostic manner.\\
\begin{figure}[t]
      	\centering
	\includegraphics[width=1\linewidth]{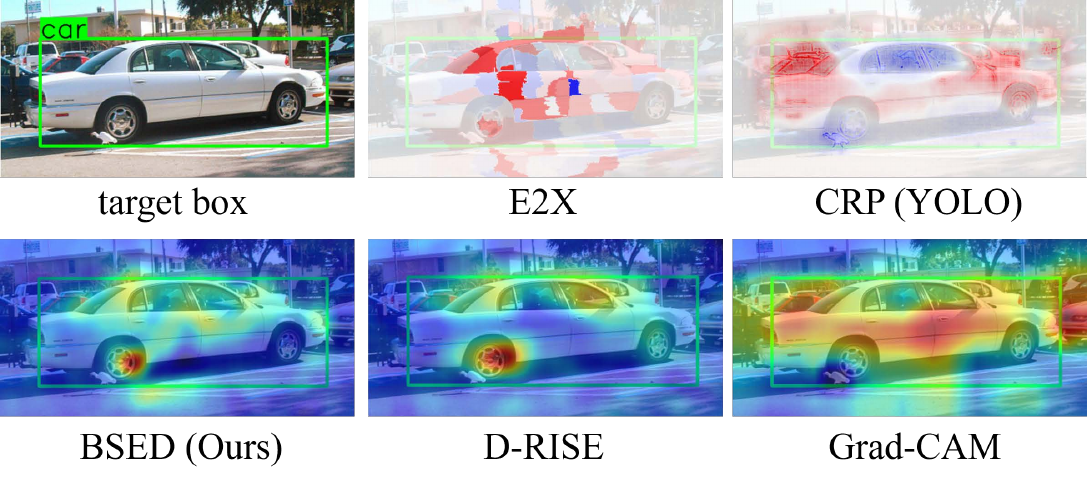}
\caption{Comparison results with existing methods in interpreting the car detection of YOLOv5s.}
\label{fig:1}
\end{figure}
\indent This study focuses on object detection tasks, such as that depicted in Fig.~\ref{fig:1}. The task requires predicting a class label and localizing a bounding box of a target object, thereby making the XAI's task of extracting information from the model more complex. Some methods extend those for image classification by adding conditions on the calculation to limit an explanation scope to a target object. Contrastive Relevance Propagation (CRP) for the YOLO detector~\cite{YOLOHeatmap} confines the calculation of attributions to those originating from nearby the target bounding box and denoting the class label associated with the target. On the other hand, D-RISE~\cite{DRISE} extends the model-agnostic approach~\cite{RISE} and calculates the feature attribution by sampling masked images and their corresponding output scores. The scores take into account both the classification and localization aspects. However, concerns remain about the explanatory validity of these methods. The imposition of the restriction on calculating attributions nearby the target object would introduce bias into the explanations. The result of CRP for the YOLO detector in Fig.~\ref{fig:1} shows that other cars within the bounding box have as high feature attributions as the target. The restriction also precludes the possibility of important clues being a little far from the target object. In addition, D-RISE is difficult to generalize to various detection targets, despite a model-agnostic method. Fig.~\ref{fig:2} illustrates the changes in saliency maps according to the parameters. A fatal error may appear in an explanation with a certain parameter, and the optimal parameter set may depend on the target, making the method difficult to apply to unknown situations. Therefore, we must pay attention to the explanatory validity. Recently, this topic has frequently been discussed with {\it axioms}~\cite{BSHAP, IG}, which refer to the properties that explainable methods should satisfy. Because the Shapley value~\cite{ShapleyValue} has been proven to satisfy the axioms, methods that extend the concept of the Shapley value to XAI have emerged to enhance the validity.\\
\indent Because research pertaining to the validity of XAI for object detection is scarce, we propose a novel method called the Baseline Shapley-based Explainable Detector (BSED). By extending Baseline Shapley~\cite{BSHAP} and applying the Shapley value to the object detection task, BSED is expected to yield explanations justified by the axioms. The technical contributions of this study are as follows.
\begin{itemize}
\item We developed an XAI for object detection that satisfies the axioms by introducing the Shapley value. We experimentally demonstrated the superior performance of our method compared to that of existing methods.
\item The inclusion of the Shapley value brings about a significant computational load. To remedy this issue, we introduced the reasonable mathematical Shapley value approximation to reduce the computation burden to a realistic cost while preserving a higher accuracy compared to existing methods.
\item We experimentally showed that our method can be applied not only in a model-agnostic manner, but also independently of fine-grained parameter tuning. This indicates that our method can interpret detection results equitably under a wide range of situations. 
\item We demonstrated the manipulation of detection results according to the explanations of our method, uncovering the potential of extending it to practical applications.
\end{itemize}
In the remainder of this paper, firstly, we summarize related studies and provide a derivation of our method. Then, we experimentally demonstrate our method's validity and possible applications. Lastly, concluding remarks are presented.

\section{Related Works}
\label{Related Works}
\begin{figure}[t]
      	\centering
      	\includegraphics[width=0.75\linewidth]{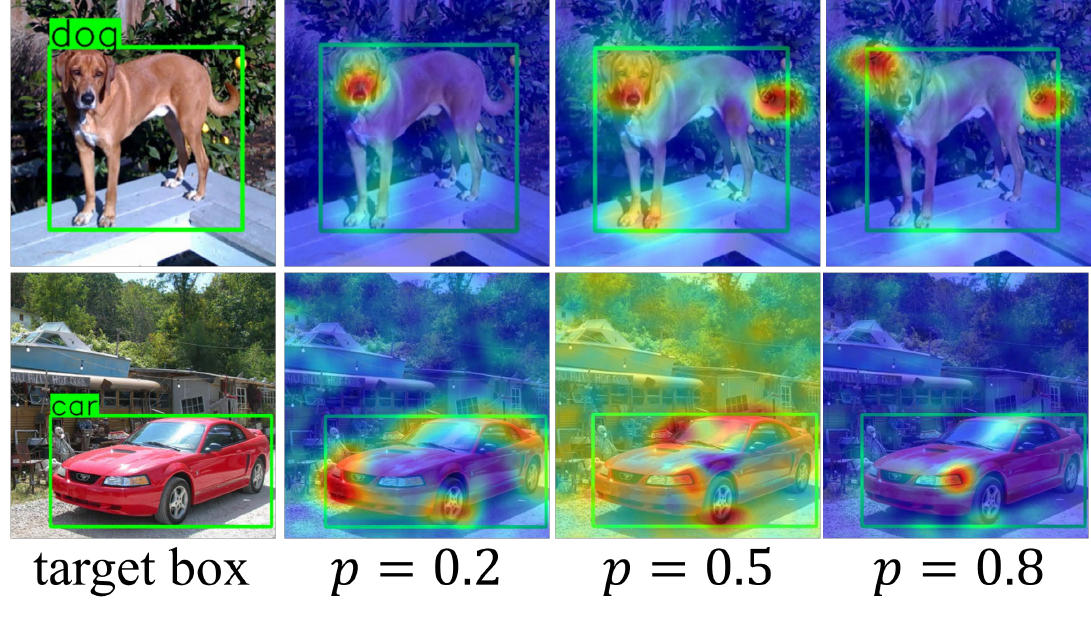}
\caption{Saliency maps generated from D-RISE explaining detection results. The parameter $p$ indicates the percentage of the non-masked area in the images for input samplings.}
\label{fig:2}
\end{figure}
\subsection{XAI for Image Classification}
Back-propagation-based methods~\cite{Gradients, SmoothGrad, DeepLIFT, GuidedBackprop} utilize gradients of the neural network to calculate pixel-wise feature attribution. For instance, Layer-wise Relevance Propagation (LRP)~\cite{LRP} involves the backward propagation of a classification score (relevance) through the neural network layers, thereby attributing relevance to each pixel in an input image. To distinguish the relevance associated with the true object from other class objects, CRP~\cite{CRP, CRPSSD} calculates the disparity between the relevance attributed to the target class and the mean relevances derived from other classes. Integrated Gradients (IG)~\cite{IG} involves varying the input image from the baseline to the original image and integrating the corresponding gradients. Activation-map-based methods~\cite{GradCAM++, ScoreCAM, SHAPCAM, CAM} utilize feature maps for the calculation. For instance, Grad-CAM~\cite{GradCAM} calculates weighted sums of the feature maps using the gradients within the neural network as the corresponding weights. These methods reduce computational complexity by extracting information from the model. However, they require prior knowledge about the optimal locations for information extraction within the models. In contrast, perturbation-based methods~\cite{ExternalPerturbation, LIME, OCCDeconv} sample perturbed inputs and the model's output for the calculation. For instance, RISE~\cite{RISE} samples partially masked images and calculates weighted sums of the input masks using the corresponding output scores as weights. Although they can generate saliency maps without knowledge about the model architecture, they are not considered suitable in cases that require real-time computation, as they sample numerous images.

\subsection{XAI for Object Detection}
Many methods~\cite{YOLOHeatmap, SSGradCAM, Revealing} extend XAI for image classification by incorporating conditions for the calculation of feature attributions that specifically target individual objects. CRP for the YOLO~\cite{YOLOHeatmap} calculates the backward propagation of a classification score. This score is associated with a region close to the target bounding box and signifies the target class label. E2X~\cite{E2X} partitions an image into superpixels and computes the average attributions across them, thereby mitigating pixel-wise noise. Although originally designed for classification, Grad-CAM can be adapted for object detection by identifying the gradients relevant to the task. D-RISE~\cite{DRISE} is an extension of a perturbation-based method called RISE~\cite{RISE} and samples binary masks for an input image. These values, 1 or 0, are determined based on probabilities $p$ and $1-p$, respectively. The output score indicates the detection similarity between detections obtained from a masked image and the target detection. A saliency map can be generated by computing the weighted sums of the input masks using the output scores as weights. Fig.~\ref{fig:1} presents a comparison of existing methods, with the explanation target being the car detection of a small YOLOv5 (YOLOv5s)~\cite{YOLOv5} model from the COCO~\cite{COCO} dataset. The methods presented in Fig.~\ref{fig:1} have been curated from diverse categories and serve as baselines for subsequent evaluations. Originally designed for classification tasks, Grad-CAM struggles to target individual objects, instead responding to all objects in the same category. E2X and CRP calculate positive and negative attributions, depicted as red and blue regions. The results of E2X contain a small quantity of noise, and are dependent on the superpixel allocation. Unlike other methods, CRP considers the car's side window as unimportant and other cars within the bounding box as important. D-RISE provides a reasonable result highlighting the important areas for car detection.

\subsection{Shapley value}
The Shapley value, originally conceptualized in cooperative game theory~\cite{ShapleyValue}, has gained prominence in the field of XAI. It offers a systematic method to distribute the ``value'' or ``contribution'' of each feature in a prediction model. However, to ensure that the interpretations provided by XAI techniques are meaningful and reliable, they must satisfy certain axioms or principles. 
 A sanity check~\cite{SanityCheck} has highlighted instances where some methods yield explanations that don't align with actual model predictions. Grounded in these concerns, various axioms have been proposed and studied in depth~\cite{IG, BSHAP, DoExp}. Here are a few key axioms:
 \vskip.5\baselineskip
\noindent {\bf Dummy.} \indent If a feature does not influence the output of the score function $f$, it can be regarded as a dummy and assigned zero attribution.
\vskip.5\baselineskip
\noindent {\bf Efficiency.} \indent The sum of the attributions across all features should equate to the difference between the output scores of the input \( \mathbf{x} \) and the baseline \( \mathbf{x^b} \), symbolically: \(\textstyle\sum\nolimits_{i=1}^{N} a_i=f(\mathbf{x})-f(\mathbf{x^b})\). This ensures that the entire contribution difference is attributed among the features.
\vskip.5\baselineskip
\noindent {\bf Linearity.} \indent For a linear combination of two score functions, \( f \) and \( g \), the attribution of the combined function should equal the sum of the attributions for each individual function.  (i.e. $a^{f+g}_i=a^{f}_i+a^{g}_i$).
\vskip.5\baselineskip
\indent Owing to its ability to satisfy these axioms, the Shapley value has been integrated into various XAIs~\cite{treeSHAP, SHAP, BSHAP, fastSHAP}. Despite its advantages, a significant drawback is the computational expense it incurs. While approximations can alleviate this computational demand, they may not always satisfy the axioms, introducing potential discrepancies in the explanations.

\section{Proposed Method}
\label{Proposed Method}
\begin{figure*}[t]
\centering
\includegraphics[width=0.75\linewidth]{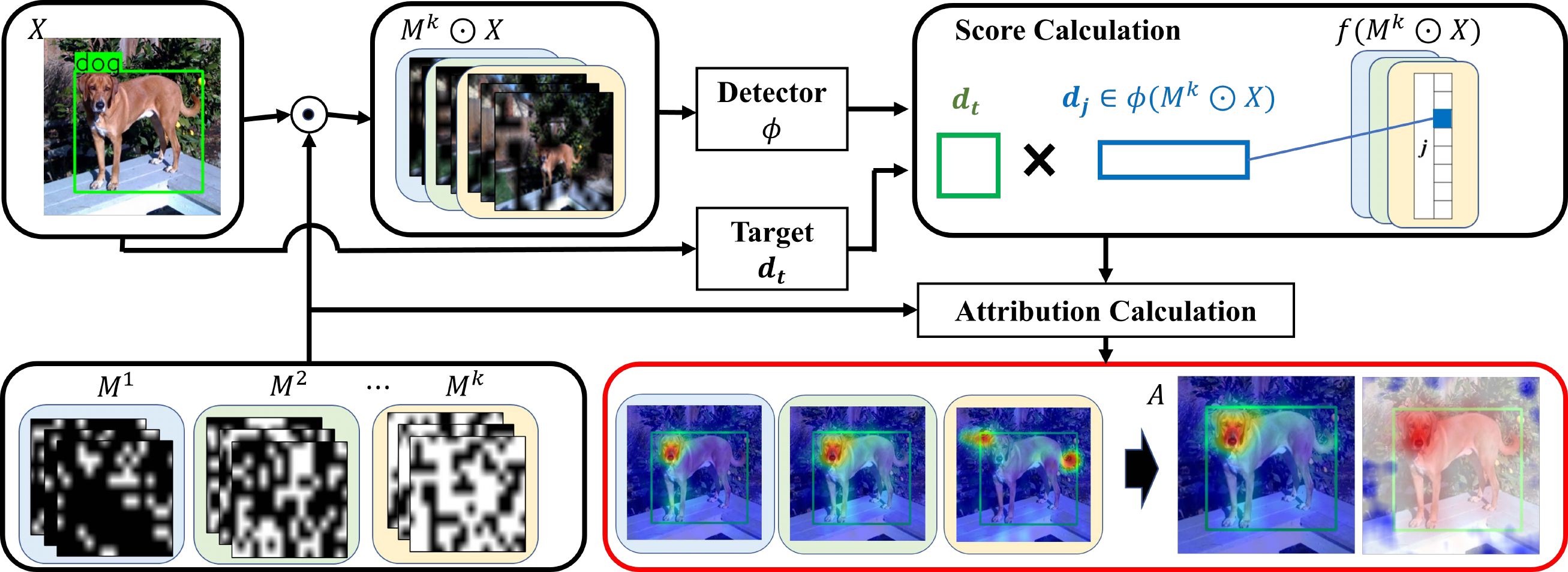}
\caption{Overview of BSED. The explanation target ${\bm d_t}$ is on the input image $X$. The detector $\phi$ obtains perturbated detections ${\bm d_j}$ from the masked image $M^k \odot X$. The similarity between the explanation target and perturbated detections is obtained to calculate the attribution on each pixel. An attribution map $A$ is generated as an explanation for obtaining the target detection.}
\label{fig:3}
\end{figure*}
\subsection{Motivation}
From the perspective of practical applicability, existing methods for XAI in object detection have two major issues. The first is the issue of validity, which has been addressed by only a few studies. Fig.~\ref{fig:2} shows that the saliency maps generated by D-RISE may vary according to parameters, and we cannot be sure whether a set of parameters is appropriate given an unknown situation. If the method is not sufficiently justified, it cannot be applied in practice. The other primary issue is that of application. Although existing methods can highlight positively important regions, they cannot sufficiently indicate regions of negative importance. In addition, their feature attributions only indicate relative importance, not clarifying how much each area contributes to the prediction. If the explanation is aimed at improving the detector's performance, the degree of impact from positive and negative areas must be important clues for deeper analysis.\\
\indent These issues have motivated us to develop an {\it axiom}-justified method that can be generalized to a wide range of situations, and provides more information for deeper analysis. In developing this method, we treated the explanation as an attribution task: the entire output score is distributed across each pixel in an image, and the attribution value indicates the contribution of each pixel to the score. Unlike existing methods, we can confirm a balance between positive and negative attributions, employing them as clues to adjust the detection results appropriately.

\subsection{Baseline Shapley}
Let us assume that $N_f$ represents all features of a model's input. Then, the Shapley value can be described as the feature attribution of the target feature $i \in N_f$ as follows:
\begin{equation}
\label{eq1}
s_i=\sum_{S\subseteq N_f\setminus i} \frac{|S|!*(|N_f|-|S|-1)!}{|N_f|!}(v(S\cup i)-v(S)).
\end{equation}
Here, $S$ denotes a subset of features excluding $i$, and $v(S)$ represents the model output when $S$ is the input. The attribution of $i$ can be obtained by averaging the marginal contributions, which are output changes by the addition of $i$. In a machine-learning setting, $v(S)$ requires retraining the model using only $S$ as an input, which is extremely time-consuming. SHAP~\cite{SHAP} avoids this process by approximating $v(S)$ as the expected value of the output scores containing $S$ as an input. However, the expected value significantly depends upon the distribution of the dataset. Therefore, we adopted another Shapley value method that is independent of the distribution of the dataset, namely Baseline Shapley~\cite{BSHAP}, which approximates $v(S)$ by combining the target input ${\bm x}$ and the baseline input ${\bm x^b}$ as follows:
\begin{equation}
\label{eq2}
v(S)=f({\bm x_S};{\bm x^b_{N_f \setminus S}}).
\end{equation}
For the input of function $f$, values corresponding to the feature set $S$ originate from ${\bm x}$, whereas those corresponding to the other $N_f \setminus S$ originate from the baseline ${\bm x^b}$. Such an application of Baseline Shapley to object detection has not been attempted previously.
\subsection{Baseline Shapley for Object Detection}
We take the function $f$ as a score function that includes an object detector, and assume the baseline $\bm x^b$ to be a black image indicating no information. We then can interpret ${\bm x_S};{\bm x^b_{N_f \setminus S}}$ as a masked image, wherein any pixels corresponding to the $S$ are the original pixels, and all other pixels are masked. If we rewrite Eq.~\ref{eq2} with the element-wise multiplication $\odot$, function to generate binary masks $\mathcal{M}(S)$, and image ${X}$, we can also reformulate Eq.~\ref{eq1} as follows:
\begin{equation}
\label{eq3}
\thickmuskip=0mu
\medmuskip=0mu
\thinmuskip=0mu
a_i=\sum_{S\subseteq N_f\setminus i}\frac{P_r(S)}{|N_f|}\Bigl\{ f\bigl( \mathcal{M}(S\cup i)\odot X\bigr)-f\bigl(\mathcal{M}(S)\odot X\bigr)\Bigr\},
\end{equation}
\begin{equation}
\label{eq4}
P_r(S)=\binom{|N_f|-1}{|S|}^{-1}.
\end{equation}
Here, we rewrite $s_i$ as $a_i$, which is the attribution value corresponding to the feature $i$. $P_r(S)$ can be represented as the reciprocal of a binomial coefficient. Subsequently, we group $S$ according to the number of non-masked pixels $|S|$.
\begin{equation}
\label{eq5}
S^l \in \Bigl\{S\subseteq {N_f\setminus i} \mathrel{\Big|} |S|=l\Bigr\} .
\end{equation}
We now can transform the summation in Eq.~\ref{eq3}.
\begin{equation}
\label{eq6}
a_i=\frac{1}{|N_f|}\sum_{l=1}^{|N_f|}F_i(S^l),
\end{equation}
\begin{dgroup*}
\begin{dmath*}
\thickmuskip=0mu
\medmuskip=0mu
\thinmuskip=0mu
F_i(S^l)=\sum_{S^l}P_r(S^l)\bigl\{ f( \mathcal{M}(S^l\cup i)\odot X)-f(\mathcal{M}(S^l)\odot X)\bigr\}
\end{dmath*}
\begin{dmath}
\label{eq7}
\thickmuskip=0mu
\medmuskip=0mu
\thinmuskip=0mu
=\mathbb{E}\bigl[ f( \mathcal{M}(S^l\cup i)\odot X)-f(\mathcal{M}(S^l)\odot X)\bigr].
\end{dmath}
\end{dgroup*}
Given that the number of $S^l$ is $\binom{|N_f|-1}{l}$, its reciprocal $P_r(S^l)$ can be considered the event probability of $S^l$. Therefore, we can approximate $F_i(S^l)$ as the expected value over $S^l$ in Eq.~\ref{eq7}. However, we realize that the calculation of Eq.~\ref{eq6} is very time-consuming, requiring $\mathcal{O}(2^{|N_f|})$ inferences to estimate the Shapley value. Therefore, we need reasonable approximations to reduce the calculation cost. 

\subsection{Approximation of Shapley value}
We can interpret Eq.~\ref{eq6} as the average of $F_i(S^l)$ over all the number of pixels. The calculation is redundant because adjacent $l$ provide similar $F_i(S^l)$. Taking this into consideration, we further reduce the computation by picking up the representative $l$ and approximating the Shapley value by $K$ layers as follows:
\begin{equation}
\label{eq8}
a_i\approx\frac{1}{K}\sum_{k=1}^{K}F_i(S^k),
\end{equation}
\begin{equation}
\label{eq9}
S^k \in \Bigl\{S\subseteq {N_f\setminus i} \mathrel{\Big|} \frac{|S|}{|N_f|}=\frac{k}{K+1}\Bigr\}.
\end{equation}
Now, $F_i(S^k)$ is the expected value of the incremental scores resulting from the participation of $i$. Inspired by the problem setting of RISE~\cite{RISE}, we simplify $F_i(S^k)$ to the expected value of incremental scores between two masks, conditioned on the event that only one of them has an element of 1 on the pixel of $i$. Here, we rewrite the mask representation from $\mathcal{M}(S^k)$ to $M^k$, which assigns binaries to all the pixels. If the pixel of $i$ has influential attribution, the contributions of the score change from the participation of $i$ should be highlighted, while those from other pixels are offset. We define the approximated score $\tilde{F}_{i,k} (\approx F_i(S^k))$ using another mask $M^{\prime k}$ as follows:
\begin{dmath}
\label{eq10}
\thickmuskip=0mu
\medmuskip=0mu
\thinmuskip=0mu
\tilde{F}_{i,k}=\mathbb{E}\bigl[{f(M^k\odot X)-f(M^{\prime k}\odot X) \mathrel{\big|} {M^k(i)-M^{\prime k}(i)=1}}\bigr].
\end{dmath}
Owing to the page limitation, details pertaining to the deformation are provided in the Appendix. We can further approximate $\tilde{F}_{i,k}$ by a simple form as follows:
\begin{dmath}
\label{eq11}
\thickmuskip=0mu
\medmuskip=0mu
\thinmuskip=0mu
\tilde{F}_{i,k} \approx \frac{\mathbb{E}\bigl[ f\bigl({M^k}\odot X\bigr)M^k(i) \bigr]-\mathbb{E}\bigl[ f\bigl({M^k}\odot X\bigr)\bigr]\cdot \mathbb{E}\bigl[M^k(i)\bigr]}{\mathbb{E}\bigl[M^k(i)\bigr] \bigl(1-\mathbb{E}\bigl[M^k(i)\bigr]\bigr)}.
\end{dmath}
Because the calculation of exact expected values is difficult, we instead apply a Monte-Carlo approximation.
\begin{align}
\label{eq12}
\mathbb{E}\bigl[M^k(i) \bigr] &\approx \frac{1}{N}\sum_{j=1}^{N} M_j^k(i), \notag \\
&=\overline{M^k(i)}.
\end{align}
Similar approximations are introduced for other expected values. In the approximation, we randomly sample $N$ binary masks. Thus, the final approximated Shapley value is
\begin{dmath}
\label{eq13}
\thickmuskip=0mu
\medmuskip=0mu
\thinmuskip=0mu
a_i\approx \frac{1}{K}\sum_{k=1}^{K}\frac{\overline{f\bigl({M^k}\odot X\bigr)M^k(i)}- \overline{f\bigl({M^k} \odot X\bigr)}\cdot \overline{M^k(i)}}{Z \cdot \overline{M^k(i)}\cdot \bigl(1-\overline{M^k(i)}\bigr)}.
\end{dmath}
Because the changes of a single pixel would have little effect on the output scores, we change pixels per patch in the mask generation. Therefore, the binary masks are initially generated in small grid size, and subsequently expand to the input image size. Thus, the contribution of the score changes should be equally distributed to all the pixels in the patch. $Z$ refers to the number of pixels in the patch and plays the role of normalization factor. The calculation of Eq.~\ref{eq13} can be performed in parallel for all pixels. Consequently, the attribution map $A$ comprising all $a_i$ is expressed as follows:
\begin{dmath}
\label{eq14}
\thickmuskip=0mu
\medmuskip=0mu
\thinmuskip=0mu
A=\frac{1}{K}\sum_{k=1}^{K}\Bigl\{ \overline{f\bigl({M^k}\odot X\bigr)M^k}- \overline{f\bigl({M^k} \odot X\bigr)}\cdot \overline{M^k} \Bigr\} \oslash \Bigl\{ Z\cdot \overline{M^k}\odot \bigl(J-\overline{M^k}\bigr)\Bigr\}.
\end{dmath}
$J$ is an all-ones matrix and $\oslash$ is an element-wise division. The number of inferences is reduced from $\mathcal{O}(2^{|N_f|})$ in Eq.~\ref{eq6} to $\mathcal{O}({N})$ in Eq.~\ref{eq14}. The overview of the process is shown in Fig.~\ref{fig:3}. Because the Shapley value can represent positive and negative attributions, we can achieve the attribution map illustrating positive areas in red and negative areas in blue. Although the definition of the term has not been clearly established, this paper refers to the saliency map as the map representing all values of attributions in the form of a heat map. In general, the introduction of approximations may degrade the accuracy of the Shapley value. We therefore conducted experiments to determine whether the attribution maps maintain high accuracy while still satisfying the axioms.

\subsection{Score Function}
The score function $f$ is inspired by the detection similarity of D-RISE~\cite{DRISE}. We define the score obtained from a masked image $M^k \odot X$ as follows:
\begin{equation}
\label{eq15}
f\bigl(M^k \odot X\bigr)=\max_{{\bm d_j} \in \phi(M^k \odot X)} \mathcal{S}({\bm d_t}, {\bm d_j}),
\end{equation}
\begin{equation}
\label{eq16}
\mathcal{S}({\bm d_t}, {\bm d_j})=s_{\rm loc}({\bm d_t}, {\bm d_j})\cdot s_{\rm cls}({\bm d_t}, {\bm d_j}).
\end{equation}
Here, $\phi$ denotes the function of the object detector, ${\bm d_j}$ indicates the vector representation of a detection result, and ${\bm d_t}$ is the vector representation of the target detection. The similarity between ${\bm d_t}$ and ${\bm d_j}$ is denoted as $\mathcal{S}({\bm d_t}, {\bm d_j})$. $s_{\rm loc}({\bm d_t}, {\bm d_j})$ is the Intersection over Union (IoU), which measures the degree of overlap between the areas of ${\bm d_t}$ and ${\bm d_j}$. $s_{\rm cls}({\bm d_t}, {\bm d_j})$ is the classification score of ${\bm d_j}$ corresponding to the class label of ${\bm d_t}$. D-RISE~\cite{DRISE} employs the cosine similarity of the class probability vectors between ${\bm d_t}$ and ${\bm d_j}$ to calculate the similarity of classification, including the probabilities for all classes. However, our method aims to calculate the attribution of the classification score pertaining to the target class. In addition, there are concerns about susceptibility to classes that are unrelated to the target. Therefore, we handle the similarity to the output class itself, rather than that of the probability vectors for all classes.
\begin{figure}[t]
      	\centering
      	\includegraphics[width=0.9\linewidth]{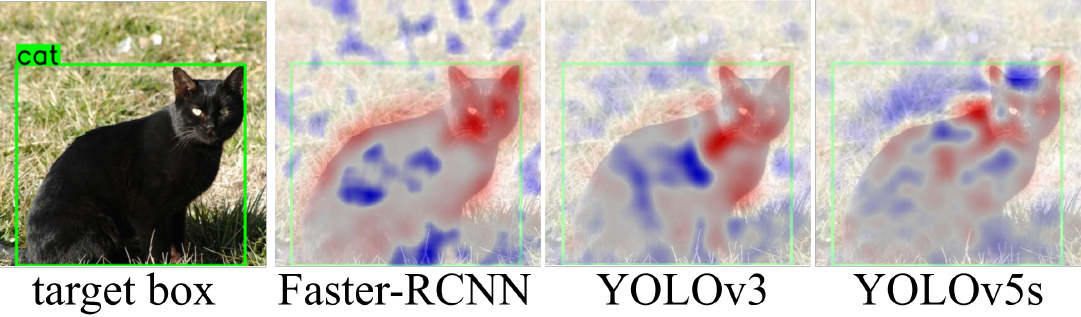}
\caption{Attribution maps on the explanation for obtaining the target detection of a cat. The explanation results from different object detectors are compared.}
\label{fig:4}
\end{figure}

\section{Evaluation and Results}
\label{Evaluation and Results}

\begin{table*}[h!t]
\centering
\small
	\begin{tabular}{|>{\centering\arraybackslash}p{6.1em}||>{\centering}p{3.3em}>{\centering}p{3.1em}>{\centering}p{3.1em}>{\centering}p{4em}>{\centering\arraybackslash}p{3.8em}|>{\centering}p{3.3em}>{\centering}p{3.1em}>{\centering}p{3.1em}>{\centering}p{4em}>{\centering\arraybackslash}p{3.8em}|}
		\hline
		& \multicolumn{5}{|c|}{COCO~\cite{COCO}} & \multicolumn{5}{c|}{VOC~\cite{VOC}}\\
		\cline{2-11}
		& EPG$(\uparrow)$ & Del.$(\downarrow)$ & Ins.$(\uparrow)$ & $\overline{\mathcal{D}(A)}(\downarrow)$ & $\overline{\mathcal{E}(A)}(\downarrow)$ & EPG$(\uparrow)$ & Del.$(\downarrow)$ & Ins.$(\uparrow)$ & $\overline{\mathcal{D}(A)}(\downarrow)$ & $\overline{\mathcal{E}(A)}(\downarrow)$ \rule[0pt]{0pt}{10pt}\\
		\hline
		Grad-CAM & 0.215 & 0.142 & 0.496 & 0.140 & - & 0.336 & 0.128 & 0.439 & 0.173 & - \\
		E2X & 0.131 & 0.094 & 0.330 & 0.016 & - & 0.239 & 0.090 & 0.273 & 0.019 & - \\
		CRP (YOLO) & 0.166 & 0.177& 0.461 & 0.008 & - & 0.260 & 0.206 & 0.359 & 0.020 & - \\
		D-RISE & 0.205 & 0.035 & 0.636 & 0.330 & - & 0.311 & 0.043 & 0.551 & 0.307 & - \\
		\hline
		BSED($K=1$) & 0.211 & 0.037 & 0.642 & 2.49${\times 10^{-6}}$ & 0.511 & 0.314 & 0.042 & 0.552 & 2.51${\times 10^{-6}}$ & 0.500 \\
		BSED($K=2$) & 0.227 & 0.034 & 0.660 & 1.87${\times 10^{-6}}$ & 0.298 & 0.328 & 0.041 & 0.571 & 1.83${\times 10^{-6}}$ & 0.286 \\
		BSED($K=4$) & {\bf 0.244} & {\bf 0.034} & {\bf 0.667} & ${\bf 1.44\times 10^{-6}}$ & {\bf 0.184} & {\bf 0.338} & {\bf 0.041} & {\bf 0.581} & ${\bf 1.41\times 10^{-6}}$& {\bf 0.176} \\
		\hline
        \end{tabular}
\caption{Results of quantitative evaluation comparison with existing methods. Energy-based Pointing Game, Deletion, and Insertion are denoted as EPG, Del., and Ins. respectively. $\overline{\mathcal{D}(A)}$ and $\overline{\mathcal{E}(A)}$ denote the metrics of {\it Dummy} and {\it Efficiency}.}
\label{tab:1}
\end{table*}

\subsection{Fundamental Evaluation}
For the calculation of BSED, the following parameters were set: patch size $Z=32\times32$, number of masks $N=6000$, and number of layers $K=4$. Unless otherwise noted, all experiments for BSED throughout this study used these parameters. Fig.~\ref{fig:1} shows that BSED effectively captures the car's characteristics, producing less noise compared to other methods. Additionally, BSED is versatile and can be applied to a range of detectors. This includes two-stage detectors, which produce region proposals before classification, and one-stage detectors that manage classification and localization simultaneously. Fig.~\ref{fig:4} presents a comparison with the target detection being the ground truth for a cat. The attribution map serves to reflect the detector's performance. Faster-RCNN~\cite{FasterRCNN}, a representative two-stage detector, can accurately detect the cat, yielding a distinct attribution map. YOLOv5s~\cite{YOLOv5}, one of the lightweight one-stage detectors, misclassified the object as a bear, failing to capture the cat's characteristics. Regarding the calculation time, BSED took 174s on a single Nvidia Tesla V100 to generate the saliency map of Fig.~\ref{fig:1}. In contrast, D-RISE required 47s for the calculation, as it does not employ a multilayer approximation. However, this approximation significantly enhances the accuracy of saliency maps. Users, especially in safety-critical fields, who seek accurate explanations might be willing to allocate more computation time and prioritize reliability over speed. Furthermore, embracing parallel computing and advancements in GPU technology can significantly cut down on computation time. As such, we don't see the processing time of BSED as a major drawback.

\subsection{Benchmark Evaluation}
\label{Benchmark Evaluation}
While various evaluation metrics have been proposed, it remains unsettled as to which ones should be consistently employed. Therefore, we selected widely employed metrics in the benchmark evaluation for a fair comparison. The Energy-based Pointing Game (EPG)~\cite{ScoreCAM} improves the traditional metrics of the Pointing Game~\cite{PG} and quantifies the degree to which attributions focus on the target object. Deletion/Insertion~\cite{InsertionDeletion} assesses the impact of pixels with high attribution values on the overall score. In Deletion, pixels with high attribution values are masked in descending order, and the degree of score reduction is evaluated. In Insertion, pixels are added to the baseline image following the same descending order, and the degree of score increase is evaluated. These evaluations were conducted using a subset of 10\% of images randomly selected from the widely used COCO~\cite{COCO} and VOC~\cite{VOC} datasets of validation splits. The detections from YOLOv5s were set as explanation targets. For a fair comparison, the score functions required for the evaluation metrics and the calculation in D-RISE are identical to that described in Eq.~\ref{eq15}. Table~\ref{tab:1} indicates BSED exhibited the best performance across all indicators. Additionally, we evaluated BSED using different numbers of layers. As the number of layers increases, the accuracy of the saliency maps likewise increases, demonstrating the efficacy of the multilayer calculation. Detailed results are provided in the Appendix.

\subsection{Evaluation for Axioms}
\begin{figure}[t]
      	\centering
      	\includegraphics[width=1\linewidth]{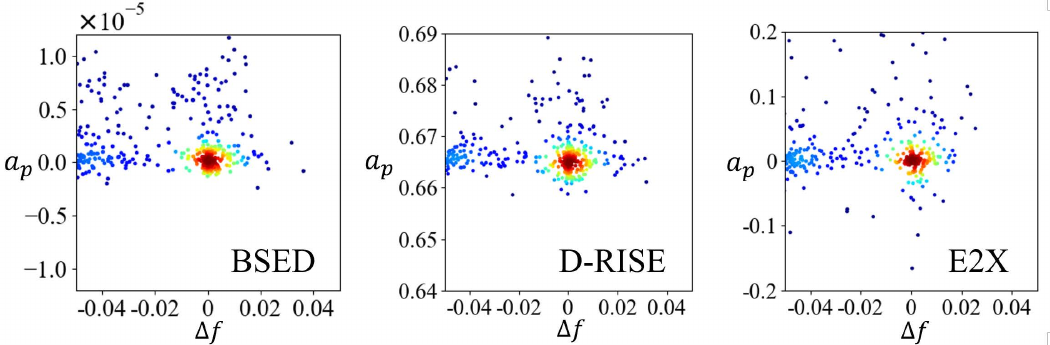}
\caption{Scatter plot showing the relationship between $\Delta f$ and $a_p$. The kernel density function was used to indicate high densities in red and low densities in blue.}
\label{fig:5}
\end{figure}

As an evaluation for {\it axioms} in XAI for object detection has not been established, we conducted new evaluations to confirm whether the method satisfies the axioms. 
\vskip.5\baselineskip
\noindent {\bf Dummy.} \indent We assessed the assignment of zero attributions for dummy features. When masking a pixel has no impact on the output score, we can interpret that pixel as a dummy feature. To make the score changes observable, we masked pixels in distinct patch regions, evaluated the change in score denoted as $\Delta f$, and determined the mean attribution values across the patch, denoted as $a_p$. The size of the patch is equivalent to the aforementioned $Z$. Fig.~\ref{fig:5} shows the relationship between these values, targeting the saliency map of Fig.~\ref{fig:1}. The patch regions were randomly created over the entire image. The plots of BSED and E2X are concentrated at the zero value, indicating that most of the dummy pixels have zero attributions. We define the criteria as follows: 
\begin{equation}
\label{eq17}
\mathcal{D}(A) = \overline{|a_d|}, \ \ a_d \in \Bigl\{ a_p \mathrel{\Big|} |\Delta f| < \sigma \Bigr\}.
\end{equation}
Here, $A$ denotes the attribution map, and $\sigma$ represents a threshold for distinguishing the dummy features. The lower value indicates the method satisfies the dummy property. We calculated the mean $\mathcal{D}(A)$ across all the attribution maps obtained in the benchmark evaluation. We set $\sigma=0.005$ and denoted the result as $\overline{\mathcal{D}(A)}$ in Table~\ref{tab:1}, showing that BSED satisfies the dummy property better than other methods.
\vskip.5\baselineskip
\noindent {\bf Efficiency.} \indent We evaluated whether the sum of the attributions is equivalent to that of the output scores of the input image. We define the efficiency metrics as follows: 
\begin{equation}
\label{eq18}
\mathcal{E}(A) = \Bigl|\sum_{a \in A}a-f(X)\Bigr|.
\end{equation}
We evaluated the average of $\mathcal{E}(A)$ over all attribution maps obtained in the benchmark evaluation. The result, as $\overline{\mathcal{E}(A)}$ in Table~\ref{tab:1}, indicates that our method most satisfies the efficiency property among the tested methods. Values for other methods are omitted since they do not meet the efficiency property, resulting in excessive values.
\vskip.5\baselineskip
\noindent {\bf Linearity.} \indent We assessed if the BSED framework satisfies linearity. If we define the similarity as $\mathcal{S}({\bm d_t}, {\bm d_j})=s_{\rm loc}({\bm d_t}, {\bm d_j})+s_{\rm cls}({\bm d_t}, {\bm d_j})$ in Eq.~\ref{eq16}, and denote the most similar detection as ${\bm d_{\max}}$, the output score $\mathcal{S}({\bm d_t}, {\bm d_{\max}})$ can be defined as a linear combination of $s_{\rm loc}({\bm d_t}, {\bm d_{\max}})$ and $s_{\rm cls}({\bm d_t}, {\bm d_{\max}})$. Because the calculation of Eq.~\ref{eq14} includes the linear calculation, the attributions must be linear combinations of those from $s_{\rm loc}({\bm d_t}, {\bm d_{\max}})$ and $s_{\rm cls}({\bm d_t}, {\bm d_{\max}})$. 

\begin{figure}[t]
      	\centering
      	\includegraphics[width=1\linewidth]{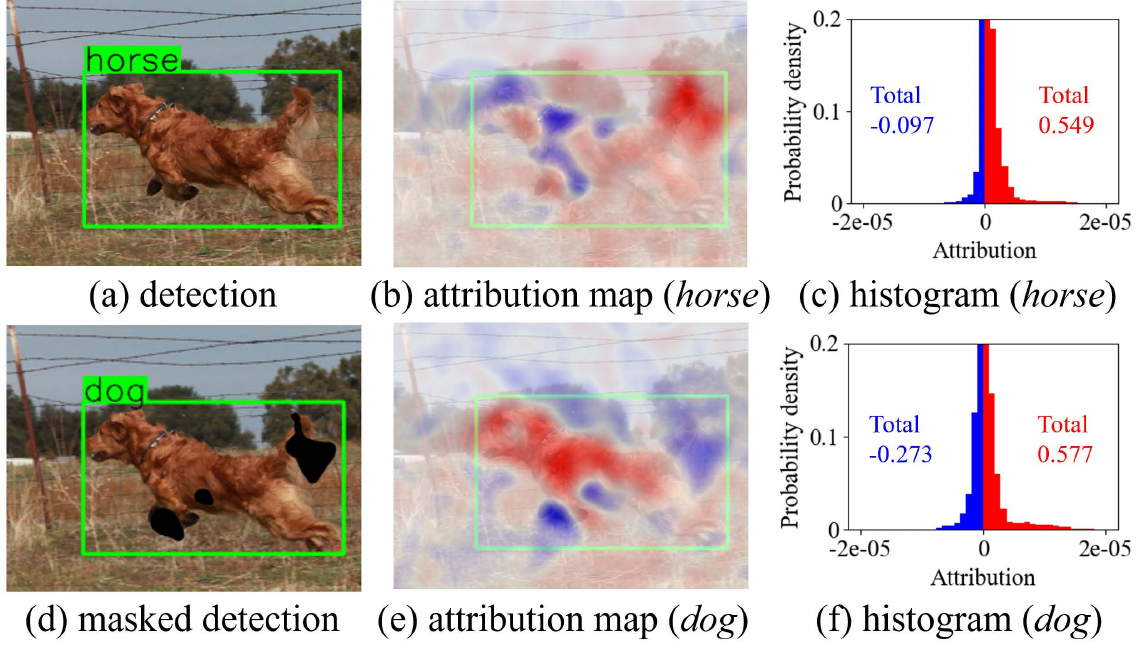}
\caption{(a) is a misclassified detection. (b) is the corresponding attribution map, and (c) is its histogram. (e) is the attribution map corresponding to a correctly classified detection, and (f) is its histogram. By masking some of the pixels of (e), a corrected detection can be obtained as shown in (d).}
\label{fig:6}
\end{figure}

\section{Discussion}
\label{Discussion}

\subsection{Analysis of False Detection}
\label{Analysis of False Detection}
YOLOv5s mistakenly identified a horse in an image of a dog, as shown in Fig.~\ref{fig:6}(a). To analyze this misclassification in depth, we generated attribution maps. By assigning the target class labels as {\it dog} and {\it horse} in Eq.~\ref{eq16}, we derived the corresponding similarity scores. Using these scores, BSED provides explanations for the two corresponding detections, as depicted in Figs.~\ref{fig:6}(b) and~\ref{fig:6}(e). The histograms and the sums of the positive and negative attribution values, respectively, are displayed in Figs.~\ref{fig:6}(c) and~\ref{fig:6}(f). We identified regions where the detector recognizes features characteristic of dogs and horses. In addition, the histogram of {\it dog} shows more negative attributions than that of {\it horse}, implying that the output score of the latter outperforms that of the former. From these results, we infer that the detector does not adequately recognize the features of a dog. The background and lower half of the body appear to contribute to misclassification. By masking pixels exhibiting negative attributions in ascending order, we can increase the classification score of {\it dog}. Fig.~\ref{fig:8} (left) shows a plot indicating the increase in the score as pixels with negative values are masked. Only a few pixels must be masked to reverse the score, thereby yielding an accurate detection shown in Fig.~\ref{fig:6}(d). 

\subsection{Analysis of True Detection}
\label{Analysis of True Detection}
\begin{figure}[t]
      	\centering
      	\includegraphics[width=1\linewidth]{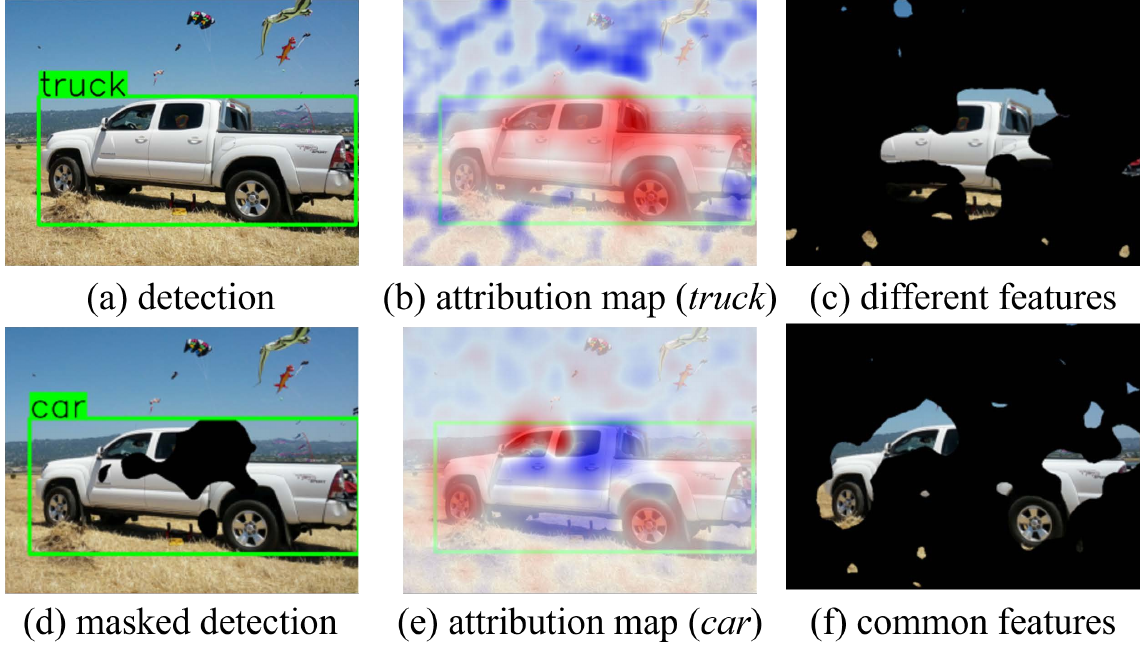}
\caption{(a) is a correct detection, and (b) is the corresponding attribution map. and (e) is the attribution map corresponding to a misclassified detection. (c) depicts the different features between {\it truck} and {\it car}, whereas (f) shows the common features between the two.}
\label{fig:7}
\end{figure}

\begin{figure}[t]
      	\centering
      	\includegraphics[width=0.86\linewidth]{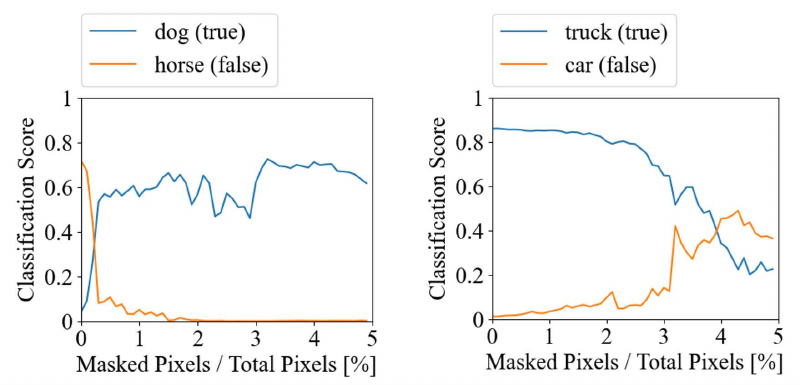}
\caption{Changes in classification score of the false detection (left) and true detection (right).}
\label{fig:8}
\end{figure}

YOLOv3~\cite{YOLOv3} accurately detected a truck, as shown in Fig.~\ref{fig:7}(a). Given the frequent confusion between trucks and cars, we examine the common features between them.  Figs.~\ref{fig:7}(b) and~\ref{fig:7}(e) show the reasoning behind the accurate and misclassified detections, respectively. Here, we regard the regions where both attributions for {\it truck} and {\it car} are positive as common features. Fig.~\ref{fig:7}(f) depicts the common features with irrelevant regions masked. These findings, suggesting indicate the tires and front parts of both {\it truck} and {\it car} are similar, are intuitive for human observers. Similarly, regions where only the attribution for {\it truck} is positive correspond to features exclusive to trucks. This is shown in Fig.~\ref{fig:7}(c), indicating that the window near the truck's bed is a specific feature for {\it truck}, which is also interpretable. Finally, an attempt is made to manipulate the true detection and achieve inaccurate classification. Masking the pixels with negative attributions for {\it car} can increase the score of {\it car}. Considering that the different features contribute only to {\it truck}, masking these features enables the detector to reduce the score of {\it truck}. Accordingly, masking both feature groups sequentially can effectively increase the score of {\it car} with a decrease in the score of {\it truck}. Fig.~\ref{fig:8} (right) shows a score inversion, resulting in a misclassification with the label {\it car}, as shown in Fig.\ref{fig:7}(d).

\section{Conclusion}
\label{Conclusion}
We proposed BSED, a promising XAI for object detection that meets the necessary criteria for explainability. BSED is not only model-agnostic but also requires no intricate parameter tuning. Our evaluations highlight BSED's superior explanatory validity compared to existing methods. We also showed that by leveraging the attributions from our method, detection results can be refined, revealing its potential for real-world applications.
 To our best knowledge, BSED is the first XAI for object detection that can be generalizable and precisely quantify both positive and negative prediction contributions. We believe this unique feature will pave the way for innovative advancements in future XAI applications.

\section*{Appendix}
\setcounter{section}{0}
\setcounter{figure}{0}
\setcounter{table}{0}
\setcounter{equation}{0}
\renewcommand{\thesection}{\Alph{section}}
\renewcommand{\thefigure}{A\arabic{figure}}
\renewcommand{\theequation}{A\arabic{equation}}
\renewcommand{\thetable}{A\arabic{table}}
\newcommand{\argmax}{\mathop{\rm arg~max}\limits}
\newcommand{\argmin}{\mathop{\rm arg~min}\limits}
\section{Advantages of the Proposed Method }
\subsection{Advantage over Existing Methods}
\begin{figure*}[tb]
\centering
\includegraphics[width=0.6\linewidth]{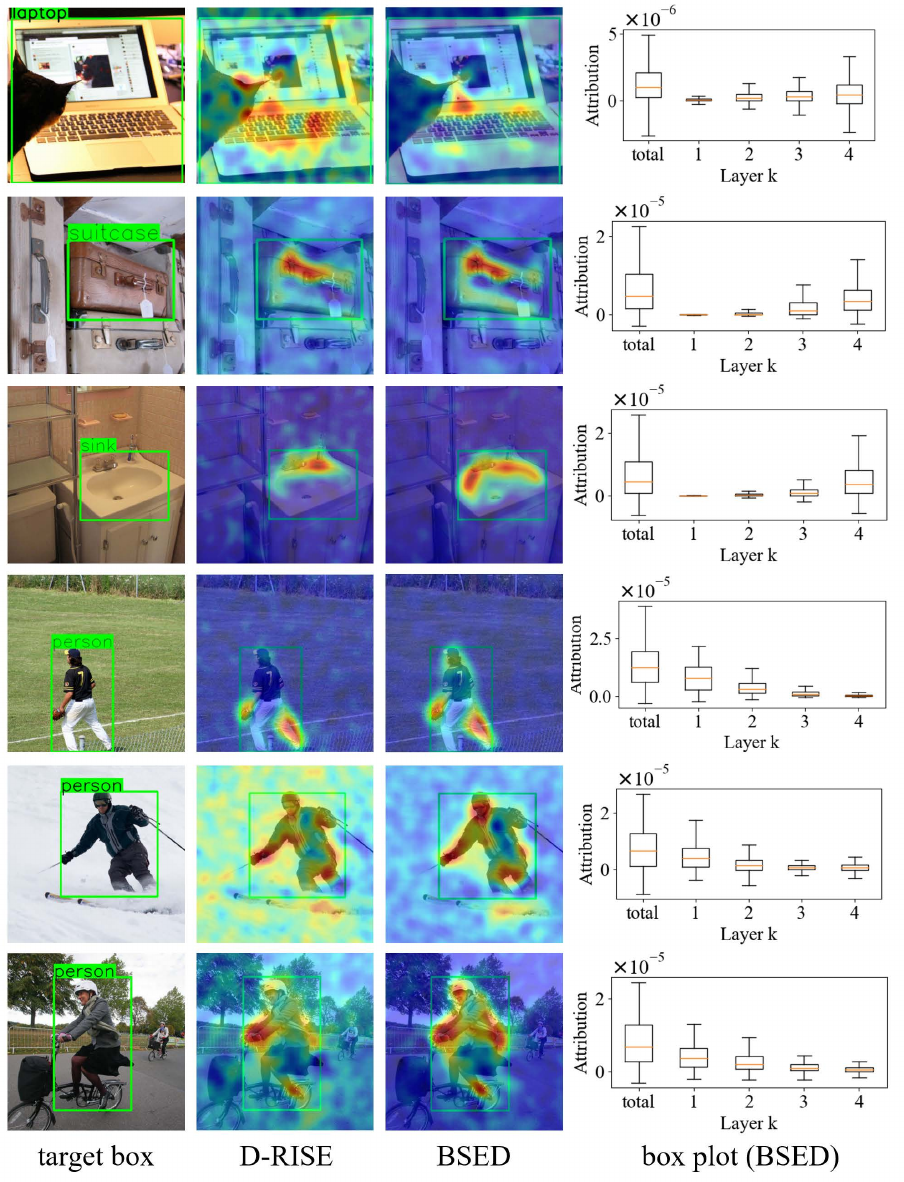}
\caption{Examples of saliency maps obtained in the benchmark evaluation of the main paper. The rightmost column shows the box plot of feature attributions calculated in each layer by BSED. In the box plot, the orange lines indicate the second interquartile, and the boxes extend from the first to the third interquartile. The topmost and bottommost points indicate the maximum and minimum attribution values, respectively.}
\label{fig:A1}
\end{figure*}

\begin{figure}[tb]
\centering
\includegraphics[width=1\linewidth]{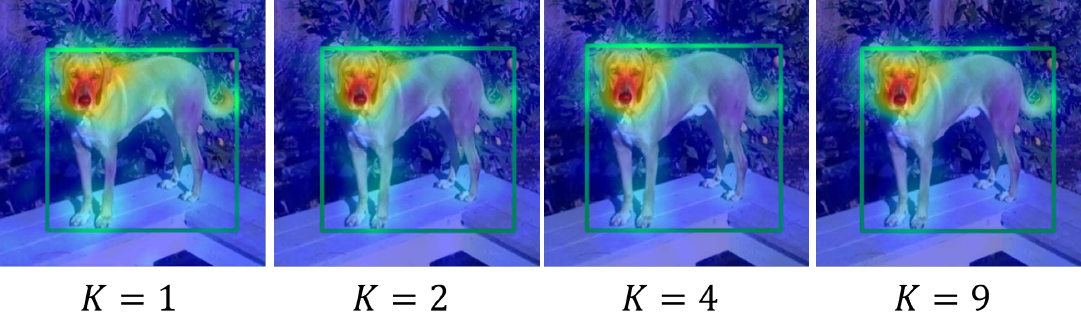}
\caption{Saliency maps targeting the same detection result as shown in Fig.~2 of the main paper, produced by varying the number of layers $K$. The saliency maps are fairly consistent and appear independent of $K$ to human eyes.}
\label{fig:A2}
\end{figure}

\begin{figure}[tb]
\centering
\includegraphics[width=1\linewidth]{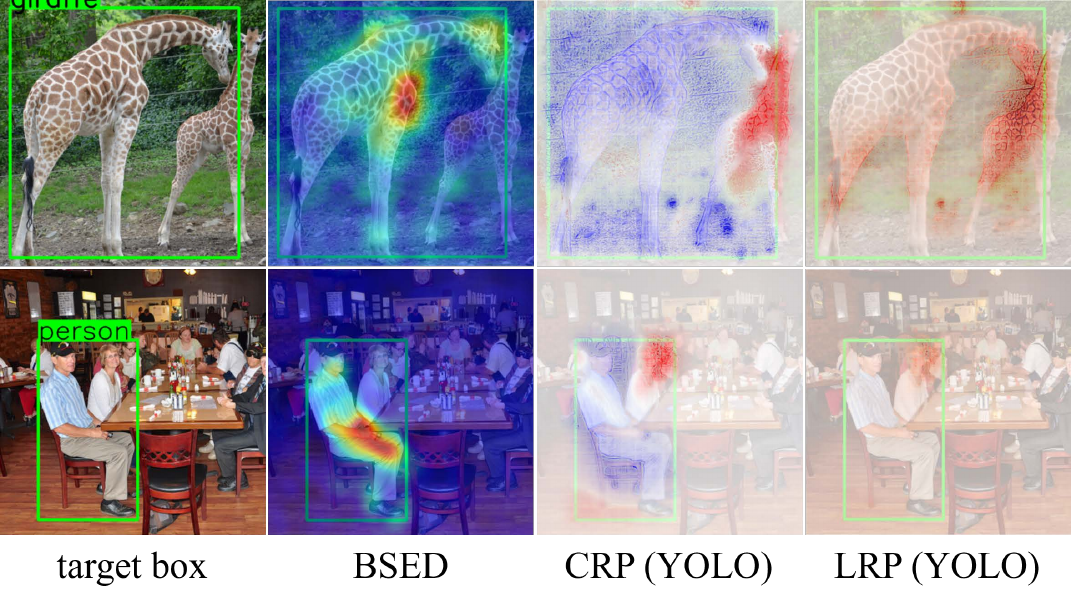}
\caption{Comparison results for detected instances containing multiple objects.}
\label{fig:comparison}
\end{figure}

In the Introduction of the main paper, concerns were highlighted about the explanatory validity of existing methods. In this section, we discuss the advantages of the proposed method for addressing each of these concerns.
\vskip.5\baselineskip
\noindent {\bf Robust Explanation.} \indent Let us clarify the advantages of our method in comparison to the existing method of D-RISE~\cite{DRISE}. It calculates weighted sums of input masks $M$ using their corresponding output scores as weights. D-RISE calculates a saliency map $A_D$ using Monte-Carlo sampling, as follows:
\begin{dmath}
\label{eq:A1}
\thickmuskip=0mu
\medmuskip=0mu
\thinmuskip=0mu
A_D=\sum_{j=1}^{N}f\bigl({M_j}\odot X\bigr)M_j.
\end{dmath}
\noindent Here, $f$, $X$, and $N$ denote the score function, an input image, and the number of masks in the sampling, respectively. $M_j$ are binary masks generated randomly, with their elements set to 1 with a probability of $p$. For a fair evaluation, our study employed the same mask generation and score function for D-RISE and BSED. In the official D-RISE study~\cite{DRISE}, the value of $p$ was set to 0.5, and we used the same value for our evaluation of D-RISE. However, generating masks using a single probability leads to inaccurate saliency maps. Fig.~\ref{fig:A1} shows examples of the saliency maps generated in the benchmark evaluation of the main paper. BSED yields interpretable saliency maps, whereas D-RISE fails to generate clear ones. We attribute this difference to the multi-layer approximation of our method. The parameter $p$ determines the proportion of masked regions in the input image in the sampling. Moreover, it significantly affects the distribution of output scores, thereby influencing the appearance of the saliency map. For instance, when the masked regions are exceedingly difficult for the target object, output scores tend to be biased toward lower values, rendering them less informative and leading to generating inaccurate saliency maps. Our method circumvents this error through multiple-layer calculations involving various ratios of masked regions. Fig.~\ref{fig:A1} shows the box plots of feature attributions calculated in each layer where the masked ratios are different. The layers which significantly affect feature attributions differ per target object, indicating that $p=0.5$ is unsuitable for certain target objects. These findings clearly show the efficacy of our multilayer calculation. In addition, Fig.~\ref{fig:A2} presents the saliency maps generated by changing the number of approximation layers~$K$. Despite the minor variation observed in the distribution of feature attributions, the explanation results remain robust to the parameter change, in contrast to results obtained by D-RISE, as shown in Fig.~2 in the main paper.
\vskip.5\baselineskip
\noindent {\bf Unconstrained Explanation.} \indent Some methods extend explainable AI (XAI) for image classification by incorporating conditions tailored for calculating feature attributions that target specific objects. For instance, Contrastive Relevance Propagation (CRP) for the YOLO detector~\cite{YOLOHeatmap} computes the backward propagation of a classification score, termed as relevance. The initial relevances associated with each point $i$ in the output layer of the YOLO network~\cite{YOLOv5} are determined as follows:
\begin{dgroup*}
\begin{dmath}
\label{eq:A2}
R_{c,b,i}=\delta_{c,i} \cdot \delta_{b,i},
\end{dmath}
\begin{dmath}
\label{eq:A3}
\delta_{c,i}=
   \begin{cases}
      p_{c,i} & \text{if $c$ is $\argmax_{c'} \{p_{c',i}\}$},\\
      0 & \text{otherwise}.
   \end{cases}
\end{dmath}
\begin{dmath}
\label{eq:A4}
\delta_{b,i}=
   \begin{cases}
      1 & \text{if $i$ is located within $b$},\\
      0 & \text{otherwise}.
   \end{cases}
\end{dmath}
\end{dgroup*}
Here, $c$ and $b$ are a class label and a bounding box of the target object, respectively. $p_{c',i}$ is the classification probability corresponding to a class label of $c'$. Besides, the initial contrastive relevance is calculated, and the difference is taken to clarify the relevance obtained from the true class object.
\begin{dgroup*}
\begin{dmath}
\label{eq:A5}
\Delta R = R_{c,b,i}-R^{cont}_{c,b,i},
\end{dmath}
\begin{dmath}
\label{eq:A6}
R^{cont}_{c,b,i}=\max_{c' \neq c} R_{c',b,i}.
\end{dmath}
\end{dgroup*}
The propagation of relevances throughout the network follows the same process as LRP~\cite{LRP}. CRP (or LRP) for the YOLO detector derives explanations by setting the initial relevances as Eq.~\ref{eq:A5} (or \ref{eq:A2}). Fig.~\ref{fig:comparison} indicates that both CRP and LRP for the YOLO detector show strong responses to other objects having the same class label within the target bounding box. The proposed method achieves equitable explanation by employing a score that simultaneously considers prediction and localization, without imposing arbitrary constraints on localization.

\clearpage
\clearpage
\subsection{Clarification of Contribution to Predictions}
\begin{figure}[tb]
\centering
\includegraphics[width=1\linewidth]{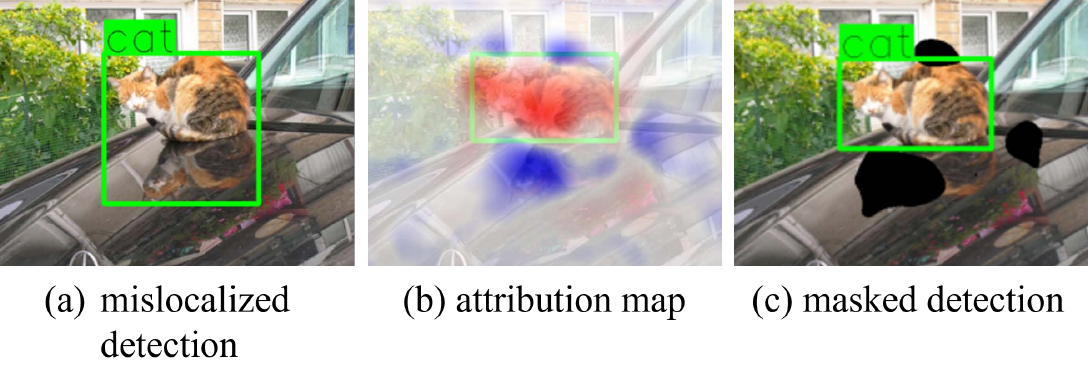}
\caption{(a) Detection of a cat with an inaccurate bounding box. (b) Attribution map on the explanation for obtaining the correctly localized detection. (c) Correct detection of the cat by masking some of the negative pixels of (b). }
\label{fig:mislocalize}
\end{figure}
\begin{figure}[tb]
\centering
\includegraphics[width=1\linewidth]{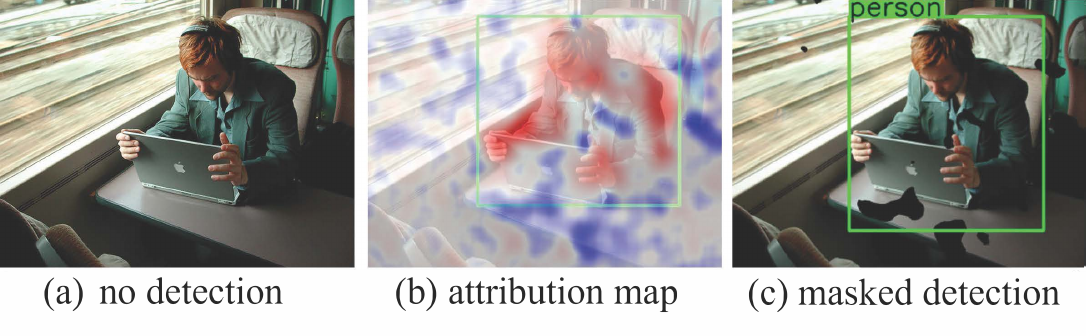}
\caption{(a) YOLOv5s fails to detect a person. (b) The attribution map illustrates the explanation for the detection attempt. (c) YOLOv5s successfully detects the person after masking specific negative pixels of (b).}
\label{fig:nodetection}
\end{figure}
\begin{figure}[tb]
\centering
\includegraphics[width=1\linewidth]{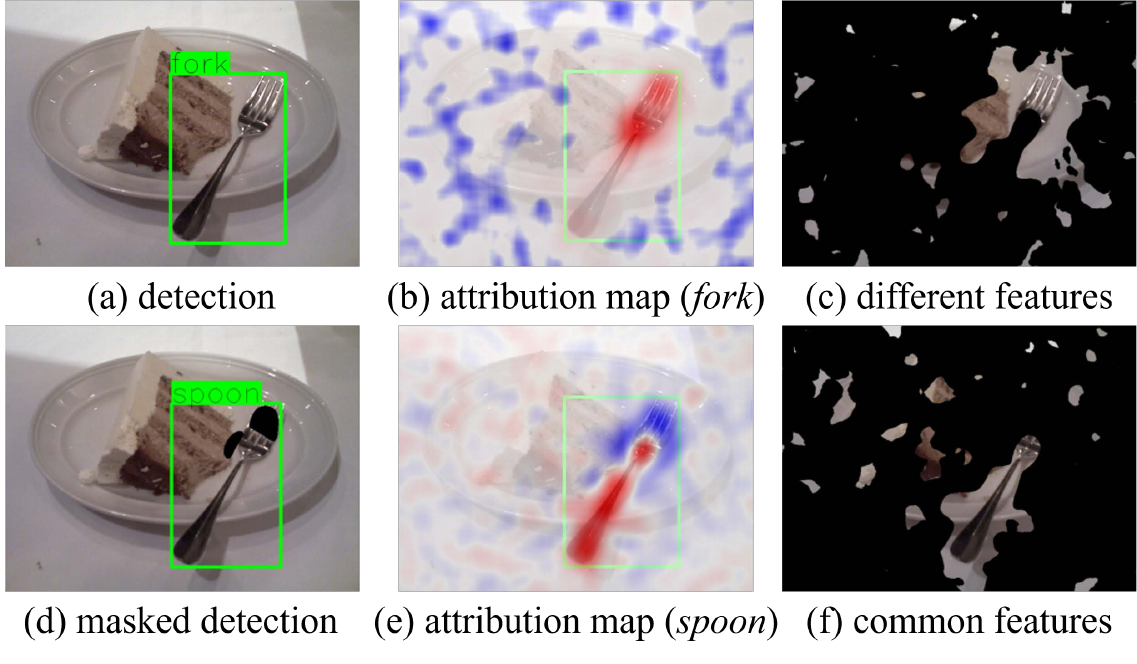}
\caption{(a) Correct detection of a fork. (b) Attribution map illustrating the rationale for the fork detection. (c) Distinctive features that only the {\it fork} possesses, setting it apart from the {\it spoon}. (d) Misclassified detection by masking specific pixels. (e) Attribution map showing the reasoning for the misclassification. (f) Common features shared between {\it fork} and {\it spoon}.}
\label{fig:truedet}
\end{figure}
Most methods that present saliency maps only highlight the regions with positive contributions to the prediction based on their unique criteria, leaving the extent of individual pixels' overall contribution unclear. The advantage of introducing the Shapley value~\cite{ShapleyValue} is that it allows for expressing the extent of contribution of each pixel to the prediction results, both positively and negatively, according to the distinct criteria. In particular, the ability to explicitly indicate negative contributions enables identifying the regions hindering accurate detection. This enables us to conduct a new analysis, such as identifying whether there is an overall lack of positive or localized negative contributions. Additionally, it allows us to evaluate if an object detector can properly distinguish the positive and negative characteristics of the target objects. In this section, we provide examples that could not be covered within the main paper.
\vskip.5\baselineskip
\noindent {\bf Mislocalized Detection.} \indent Fig.~\ref{fig:mislocalize} shows an example of a mislocalized detection. YOLOv5s~\cite{YOLOv5} detected a cat with an inaccurate bounding box encompassing reflections from a nearby car, as shown in Fig.~\ref{fig:mislocalize}(a). By setting the cat's ground truth as the target detection, we can generate an attribution map on the explanation for obtaining the correct detection, as shown in Fig.~\ref{fig:mislocalize}(b). Notably, the negative attributions observed on the cat's face were projected onto the car, contributing to the mislocalization. By masking the pixels with negative attributions in ascending order, the detection similarity to the ground truth of the cat can be increased. With a few pixels masked, YOLOv5s can detect the cat with an accurate bounding box, as shown in Fig.~\ref{fig:mislocalize}(c).
\vskip.5\baselineskip
\noindent {\bf No Detection.} \indent Fig.~\ref{fig:nodetection} shows an example of a no detection. YOLOv5s~\cite{YOLOv5} could not detect a person as shown in Fig.~\ref{fig:nodetection}(a). By setting the ground truth of the person as the target detection, we can create an attribution map on the explanation for obtaining the correct detection, as shown in Fig.~\ref{fig:nodetection}(b). Negative attributions are gathered on the background, and they contribute to the decrease in the classification score of {\it person}. By masking the pixels with negative attributions in ascending order, the detection similarity to the ground truth was increased. With a few pixels masked, YOLOv5s can correctly detect the person, as shown in Fig.~\ref{fig:nodetection}(c).
\vskip.5\baselineskip
\noindent {\bf Additional Example of True Detection.} \indent Fig.~\ref{fig:truedet}(a) shows that YOLOv3~\cite{YOLOv3} correctly detected a fork. We generated attribution maps, as depicted in Figs.~\ref{fig:truedet}(b) and~\ref{fig:truedet}(e), following the same procedure as in the main paper. We illustrated different and common features between {\it fork} and {\it spoon}, as shown in Figs.~\ref{fig:truedet}(c) and~\ref{fig:truedet}(f). By masking the pixels with negative attributions for {\it spoon} and positive attributions for {\it fork}, YOLOv3 can obtain a misclassified detection with the class of {\it spoon}, as shown in Fig.~\ref{fig:truedet}(d).\\

\section{Details of Benchmark Evaluation}
\subsection{Parameter Configuration}
In the benchmark evaluation of the main paper, we set the following parameters: patch size $Z = 32 \times 32$, number of masks $N = 6000$, and number of layers $K = 4$. The size of all input images employed in this study were resized to approximately 600 pixels to fully leverage the performance of the object detectors. The patch size $Z$, which affects the resolution of the attribution maps, depends upon the size of the input image. The smaller $Z$ provides saliency maps with more speckles because high-resolution masks require a larger number of samplings for a good shapley value approximation.  We empirically selected this parameter through a few experiments. The values of $N$ and $K$ affect the approximation accuracy of Eq.~14 in the main paper. We consider $N=6000$ reasonable, as other explainable methods~\cite{RISE, DRISE, OCCAM} using Monte Carlo sampling tend to employ parameters of approximately the same magnitude. Therefore, the computational cost of BSED increases in proportion to~$K$. The result of Fig.~\ref{fig:A2} shows that $K = 4$ is sufficiently accurate for human eyes while refraining from the increase in calculation costs.

\subsection{Environmental Setting}
The GPU used in our evaluation was a single Tesla V100 model with 32GB of memory. The software environment consisted of the Ubuntu 18.04 operating system. Detailed information about the libraries, package dependencies, and pre-trained object detectors used for the evaluation are provided in the appendix for the source code.

\subsection{Metrics}
\begin{figure}[tb]
\centering
\includegraphics[width=1\linewidth]{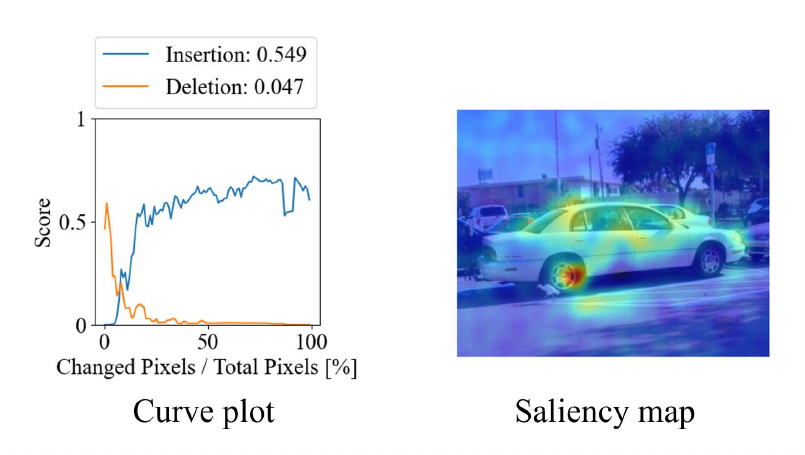}
\caption{Curve plot for the Deletion and Insertion evaluation (left) derived from the saliency map (right). The horizontal axis represents the percentage of the number of removed or added pixels. The AUC of these plots is also shown.}
\label{fig:A7}
\end{figure}

\begin{figure}[tb]
\centering
\includegraphics[width=1\linewidth]{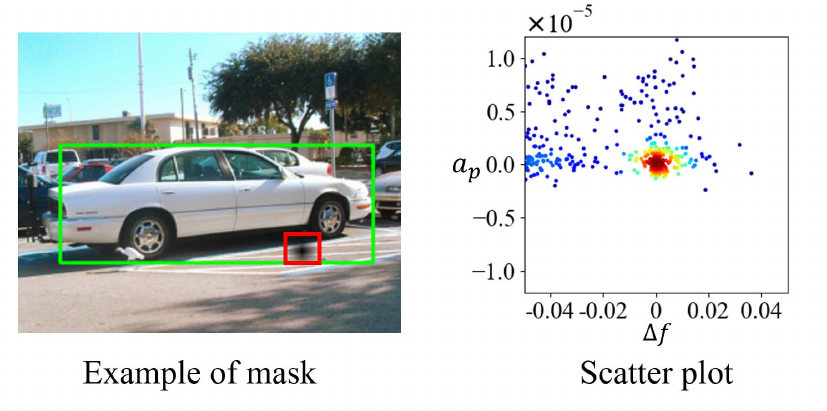}
\caption{Example of a mask applied to the input image in the Dummy evaluation. The red rectangle highlights the location where the mask was applied (left). Scatter plot illustrates the relationship between $\Delta f$ and $a_p$ (right).}
\label{fig:A8}
\end{figure}

\noindent {\bf Energy-based Pointing Game.} \indent Pointing Game~\cite{PG} has been widely-used evaluation metric in various research, which measures whether the pixel with the highest attribution is within the target bounding box. However, this method does not consider the attribution of other pixels. Therefore, we adopted Energy-based Pointing Game~\cite{ScoreCAM} to consider the states of other attributions. In this metric, the number of feature attributions gathered in the target bounding box is evaluated as follows:
\begin{dmath}
\label{eq:A7}
\thickmuskip=0mu
\medmuskip=0mu
\thinmuskip=0mu
\frac{\sum{}L_{(i,j)\in bbox}}{\sum{}L_{(i,j)\in bbox}+\sum{}L_{(i,j)\notin bbox}}.
\end{dmath}
Here, $L_{(i,j)\in bbox}$ denotes the attribution value of any pixel $(i, j)$ which is located inside the target bounding box. Because some methods compared in the benchmark evaluation calculated negative feature attributions, we normalized the attributions by the min-max normalization for the calculation of Eq.~\ref{eq:A7}.
\vskip.5\baselineskip
\noindent {\bf Deletion and Insertion.} \indent We adopted deletion and insertion~\cite{InsertionDeletion} for the benchmark evaluation of the main paper because these are also widely adopted by various research. In the deletion metric, pixels with higher attributions are sequentially removed from the input image, and the corresponding decrease in the output score of the model is evaluated. In the insertion metric, pixels are added to the black image in the same order, and the corresponding increase in the output score of the model is evaluated. Fig.~\ref{fig:A7} shows the curve plots of these metrics calculated from the saliency map shown to its right. The area under the curve (AUC) is an indicator of this quantitative evaluation. 
\vskip.5\baselineskip
\noindent {\bf Dummy evaluation.} \indent In the evaluation of the dummy property, we generated masks for patch areas of the input image and evaluated the change of score $\Delta f$ and the average of attribution values within the patch $a_p$. The patch regions are randomly created over the entire image. Fig.~\ref{fig:A8} shows an example of a mask applied to a patch area in the input image, as indicated by the red rectangle.

\begin{table}[t]
\setlength{\tabcolsep}{2pt}
\centering
\small
\begin{tabular}{|c||ccccc|}
	\hline
	& \multicolumn{5}{|c|}{YOLOv3~\cite{YOLOv3}}\\
	\cline{2-6}
	& EPG$(\uparrow)$ & Del.$(\downarrow)$ & Ins.$(\uparrow)$ & $\overline{\mathcal{D}(A)}(\downarrow)$ & $\overline{\mathcal{E}(A)}(\downarrow)$ \rule[0pt]{0pt}{10pt} \\
	\hline
	D-RISE & 0.219 & 0.053 & 0.737 & 0.475 & - \\
	BSED (K=1) & 0.225 & 0.055 & 0.739 & $2.98\times 10^{-6}$ & 0.569 \\
	BSED (K=2) & 0.247 & 0.053 & 0.762 & $2.40\times 10^{-6}$ & 0.324 \\
	BSED (K=4) & {\bf 0.269} & {\bf 0.052} & {\bf 0.770} & ${\bf 1.94\times 10^{-6}}$ & {\bf 0.215} \\
	\hline
\end{tabular}
\begin{tabular}{|c||ccccc|}
	\hline
	& \multicolumn{5}{|c|}{Faster-RCNN~\cite{FasterRCNN}}\\
	\cline{2-6}
	& EPG$(\uparrow)$ & Del.$(\downarrow)$ & Ins.$(\uparrow)$ & $\overline{\mathcal{D}(A)}(\downarrow)$ & $\overline{\mathcal{E}(A)}(\downarrow)$ \rule[0pt]{0pt}{10pt} \\
	\hline
	D-RISE & 0.275 & 0.213 & 0.826 & 0.563 & - \\
	BSED (K=1) & 0.279 & 0.214 & 0.826 & $2.41\times 10^{-6}$ & 0.573 \\
	BSED (K=2) & 0.304 & 0.204 & 0.843 & $1.84\times 10^{-6}$ & 0.325 \\
	BSED (K=4) & {\bf 0.327} & {\bf 0.201} & {\bf 0.853} & ${\bf 1.43\times 10^{-6}}$ & {\bf 0.182} \\
	\hline
\end{tabular}
\caption{Results of quantitative evaluation using the same metrics as in the benchmark evaluation. The explanations of detection results derived from the COCO~\cite{COCO} dataset were evaluated.}
\label{tab:A1}
\end{table}

\subsection{Discussion}
\noindent {\bf Evaluation Results.} \indent Our proposed methods outperform the existing methods evaluated in the benchmark evaluation. Nevertheless, in some metrics, we observed the performance of D-RISE~\cite{DRISE} is on par with BSED when the number of approximation layers $K$ is set to $1$. This resemblance can be attributed to the algorithms employed by both methods. According to Eq.~14, we notice the computation of BSED $(K=1)$ contains a similar term as $A_D$ in Eq.~\ref{eq:A1}. When the mask generation and score functions are identical to those of D-RISE, the saliency map of BSED $(K=1)$ can be derived through subtraction and division of uniform values for each element of $A_D$. These operations can be performed without altering the order of the feature attribution magnitudes. Because deletion and insertion~\cite{InsertionDeletion} depend on the order of the feature attribution magnitudes, the evaluation results are similar between the two methods.\\ \indent In the benchmark evaluation, we employed novel metrics for {\it axioms}, as well as metrics commonly used in other papers. However, the paradigm for evaluating explanations aligns with the breakthroughs in XAI and will continue to evolve in the future. While this paper makes a significant contribution by introducing a novel method, establishing a robust framework for assessing the explanatory validity remains a task for future works.
\vskip.5\baselineskip
\noindent {\bf Evaluation with Other Object Detectors.} \indent We conducted a benchmark evaluation using target detections from other object detectors, YOLOv3~\cite{YOLOv3} and Faster-RCNN~\cite{FasterRCNN}. Table~\ref{tab:A1} indicates that our method maintains the same superior performance observed in the benchmark evaluation from the main paper. We can also confirm the efficacy of the multilayer approximation. These results quantitatively indicate the proposed method can be applied to various detectors in a model-agnostic manner.
\vskip.5\baselineskip
\noindent {\bf Significant Difference.} \indent To confirm significant differences, we conducted a one-sided test between D-RISE and BSED ($K=4$) for evaluation results in our study. P-values are shown in Table~\ref{tab:A2}. Our BSED is statistically significant ($p<0.05$) in most cases. For the deletion metrics, removing only a few critical pixels significantly drops the scores and makes the evaluation value, which is the area under the curve shown in Fig.~\ref{fig:A7}, close to zero. This makes the differences between the results of the two methods less noticeable.

\begin{table}[t]
\setlength{\tabcolsep}{2pt}
\centering
\small
	\begin{tabular}{|>{\centering}p{4.7em}>{\centering}p{3.8em}||>{\centering}p{2.8em}>{\centering}p{2.8em}>{\centering}p{2.8em}>{\centering}p{2.8em}>{\centering\arraybackslash}p{3.4em}|}
		\hline
		& & \multicolumn{5}{|c|}{P-Value (one-sided)} \\
		\hline
		Detector & Dataset & E.P. & Del. & Ins. & $\overline{\mathcal{D}(A)}$ & $\overline{\mathcal{E}(A)}$ \rule[0pt]{0pt}{10pt} \\
		\hline
		YOLOv5s & COCO & $<$0.001 & 0.361 & $<$0.001 & $<$0.001 & $<$0.001 \\
		YOLOv5s & VOC & $<$0.001 & 0.040 & $<$0.001 & $<$0.001 & $<$0.001 \\
		YOLOv3 & COCO & $<$0.001 & 0.172 & $<$0.001 & $<$0.001 & $<$0.001 \\
		FasterRCNN & COCO & $<$0.001 & $<$0.001 & $<$0.001 & $<$0.001 & $<$0.001 \\
		\hline
        \end{tabular}
\caption{One-sided test for evaluation results conducted in our study.}
\label{tab:A2}
\end{table}

\onecolumn
\section{Details of Algorithm}
\subsection{Transformation of Equation}
In this section, we describe the details of the transformation from Eq. 10 to Eq. 11 in the main paper. Note that the equation numbers below correspond to those in the main paper.

\begin{dmath}[number={10}]
\thickmuskip=0mu
\medmuskip=0mu
\thinmuskip=0mu
\tilde{F}_{i,k}=\mathbb{E}\bigl[{f(M^k\odot X)-f(M^{\prime k}\odot X) \mathrel{\big|} {M^k(i)-M^{\prime k}(i)=1}}\bigr].
\end{dmath}

\noindent We can express the expected value of Eq.~\ref{eq10} as the summation of all combinations of two mask patterns. We describe two binary masks as $M^k_a$ and $M^k_b$, which have similar patterns to $M^k$. If they are allowed to be duplicated, we should consider two conditions between the masks, namely $M^k_a(i)-M^k_b(i)=1$ and $M^k_b(i)-M^k_a(i)=1$.
\begin{dmath}[number={10a}]
\label{eq10a}
\thickmuskip=0mu
\medmuskip=0mu
\thinmuskip=0mu
\tilde{F}_{i,k}=\sum_{m^k_a}\sum_{m^k_b}\biggl\{ \Bigl(f(m^k_a\odot X)-f(m^k_b\odot X)\Bigr)P\Bigl[{M^k_a=m^k_a}, {M^k_b=m^k_b} \mathrel{\Big|} {M^k_a(i)-M^k_b(i)=1}\Bigr] + \Bigl(f(m^k_b\odot X)-f(m^k_a\odot X)\Bigr)P\Bigl[{M^k_a=m^k_a}, {M^k_b=m^k_b} \mathrel{\Big|} {M^k_b(i)-M^k_a(i)=1}\Bigr]\biggr\}.
\end{dmath}

\noindent Here, $P$ indicates the probability. This equation can be further transformed as follows:

\begin{dgroup*}
\begin{dmath}[number={10b}]
\label{eq10b}
\thickmuskip=0mu
\medmuskip=0mu
\thinmuskip=0mu
\tilde{F}_{i,k}=\sum_{m^k_a}\sum_{m^k_b}\Biggl\{ \frac{\bigl(f(m^k_a\odot X)-f(m^k_b\odot X)\bigr)P\bigl[{M^k_a=m^k_a}, {M^k_b=m^k_b}, {M^k_a(i)-M^k_b(i)=1}\bigr]}{P[M^k_a(i)-M^k_b(i)=1]}+ \frac{\bigl(f(m^k_b\odot X)-f(m^k_a\odot X)\bigr)P\bigl[{M^k_a=m^k_a}, {M^k_b=m^k_b}, {M^k_b(i)-M^k_a(i)=1}\bigr]}{P[M^k_b(i)-M^k_a(i)=1]} \Biggr\}.
\end{dmath}
\begin{dmath}[number={10c}]
\label{eq10c}
\thickmuskip=0mu
\medmuskip=0mu
\thinmuskip=0mu
=\sum_{m^k_a}\sum_{m^k_b}\Biggl\{ \frac{\bigl(f(m^k_a\odot X)-f(m^k_b\odot X)\bigr)P\bigl[{M^k_a=m^k_a}, {M^k_b=m^k_b}, {M^k_a(i)-M^k_b(i)=1}\bigr]}{P[M^k_a(i)=1]\cdot P[M^k_b(i)=0]}+ \frac{\bigl(f(m^k_b\odot X)-f(m^k_a\odot X)\bigr)P\bigl[{M^k_a=m^k_a}, {M^k_b=m^k_b}, {M^k_b(i)-M^k_a(i)=1}\bigr]}{P[M^k_b(i)=1]\cdot P[M^k_a(i)=0]} \Biggr\}.
\end{dmath}
\begin{dmath}[number={10d}]
\label{eq10d}
\thickmuskip=0mu
\medmuskip=0mu
\thinmuskip=0mu
=\frac{1}{P[M^k(i)=1]\cdot P[M^k(i)=0]}\sum_{m^k_a}\sum_{m^k_b}\biggl\{ \bigl(f(m^k_a\odot X)-f(m^k_b\odot X)\bigr)P\bigl[{M^k_a=m^k_a}, {M^k_b=m^k_b}, {M^k_a(i)-M^k_b(i)=1}\bigr] + \bigl(f(m^k_b\odot X)-f(m^k_a\odot X)\bigr)P\bigl[{M^k_a=m^k_a}, {M^k_b=m^k_b}, {M^k_b(i)-M^k_a(i)=1}\bigr]\biggr\}.
\end{dmath}
\end{dgroup*}

\noindent We can divide the patterns by the combination of $m^k_a(i)$ and $m^k_b(i)$. 

\begin{dmath}[number={10e}]
\label{eq10e}
\thickmuskip=0mu
\medmuskip=0mu
\thinmuskip=0mu
P\Bigl[{M^k_a=m^k_a}, {M^k_b=m^k_b}, {M^k_a(i)-M^k_b(i)=1}\Bigr] =
   \begin{cases}
      P\Bigl[{M^k_a=m^k_a}, {M^k_b=m^k_b}\Bigr] & \text{if $m^k_a(i)-m^k_b(i)=1$},\\
      0 & \text{if $m^k_a(i)-m^k_b(i)=0$},\\
      0 & \text{if $m^k_a(i)-m^k_b(i)=-1$}.
   \end{cases}
\end{dmath}
\begin{dmath}[number={10f}]
\label{eq10f}
\thickmuskip=0mu
\medmuskip=0mu
\thinmuskip=0mu
P\Bigl[{M^k_a=m^k_a}, {M^k_b=m^k_b}, {M^k_b(i)-M^k_a(i)=1}\Bigr] =
   \begin{cases}
      P\Bigl[{M^k_a=m^k_a}, {M^k_b=m^k_b}\Bigr] & \text{if $m^k_b(i)-m^k_a(i)=1$},\\
      0 & \text{if $m^k_b(i)-m^k_a(i)=0$},\\
      0 & \text{if $m^k_b(i)-m^k_a(i)=-1$}.
   \end{cases}
\end{dmath}

\noindent Subsequently , we can reformulate Eq.~\ref{eq10d} as follows: 

\begin{dmath}[number={10g}]
\label{eq10g}
\thickmuskip=0mu
\medmuskip=0mu
\thinmuskip=0mu
\tilde{F}_{i,k}
=\frac{1}{P[M^k(i)=1]\cdot P[M^k(i)=0]}\sum_{m^k_a}\sum_{m^k_b}\Bigl\{ \Bigl(f(m^k_a\odot X)-f(m^k_b\odot X)\Bigr)\Bigl(m^k_a(i)-m^k_b(i)\Bigr)P\Bigl[{M^k_a=m^k_a}, {M^k_b=m^k_b}\Bigr]\Bigr\}.
\end{dmath}

\noindent  We now seek to reformulate the summation of $m^k_b$ into its expected value.
\begin{dgroup*}
\begin{dmath*}
\thickmuskip=0mu
\medmuskip=0mu
\thinmuskip=0mu
\sum_{m^k_a}\sum_{m^k_b}\Bigl\{ \Bigl(f(m^k_a\odot X)-f(m^k_b\odot X)\Bigr)\Bigl(m^k_a(i)-m^k_b(i)\Bigr)P\Bigl[{M^k_a=m^k_a}, {M^k_b=m^k_b}\Bigr]\Bigr\}
\end{dmath*}
\begin{dmath}[number={10h}]
\label{eq10h}
\thickmuskip=0mu
\medmuskip=0mu
\thinmuskip=0mu
=\sum_{m^k_a}\Bigl\{f(m^k_a\odot X)\cdot m^k_a(i)-f(m^k_a\odot X)\cdot \mathbb{E}\bigl[M^k_b(i)\bigr]-\mathbb{E}\bigl[f(M^k_b\odot X)\bigr]\cdot m^k_a(i)+\mathbb{E}\bigl[f(M^k_b\odot X)\cdot M^k_b(i) \bigr]  \Bigr\}P\Bigl[{M^k_a=m^k_a}\Bigr]
\end{dmath}
\begin{dmath}[number={10i}]
\label{eq10i}
\thickmuskip=0mu
\medmuskip=0mu
\thinmuskip=0mu
\approx \sum_{m^k_a}\Bigl\{f(m^k_a\odot X)\cdot m^k_a(i)-f(m^k_a\odot X)\cdot \mathbb{E}\bigl[M^k_b(i)\bigr]-\mathbb{E}\bigl[f(M^k_b\odot X)\bigr]\cdot m^k_a(i)+\mathbb{E}\bigl[f(M^k_b\odot X)\bigr]\cdot \mathbb{E}\bigl[M^k_b(i)\bigr] \Bigr\}P\Bigl[{M^k_a=m^k_a}\Bigr]
\end{dmath}
\begin{dmath}[number={10j}]
\label{eq10j}
\thickmuskip=0mu
\medmuskip=0mu
\thinmuskip=0mu
=\sum_{m^k_a}\Bigl\{\Bigl(f(m^k_a\odot X)-\mathbb{E}\bigl[f(M^k_b\odot X)\bigr] \Bigr)\Bigl(m^k_a(i)-\mathbb{E}\bigl[M^k_b(i)\bigr] \Bigr)\Bigr\}P\Bigl[{M^k_a=m^k_a}\Bigr].
\end{dmath}
\end{dgroup*}

\noindent In the transformation, we assumed the independence between $f(M^k_b\odot X)$ and $M^k_b(i)$. Given that $m^k_a$ and $m^k_b$ follow the same distribution of $M^k$,  we can rewrite Eq.~\ref{eq10g} as follows.

\begin{dgroup*}
\begin{dmath}[number={10k}]
\label{eq10k}
\thickmuskip=0mu
\medmuskip=0mu
\thinmuskip=0mu
\tilde{F}_{i,k}
\approx \frac{1}{P[M^k(i)=1]\cdot P[M^k(i)=0]}\sum_{m^k}\Bigl\{\Bigl(f(m^k\odot X)-\mathbb{E}\bigl[f(M^k\odot X)\bigr] \Bigr)\Bigl(m^k(i)-\mathbb{E}\bigl[M^k(i)\bigr] \Bigr)\Bigr\}P\Bigl[{M^k=m^k}\Bigr]
\end{dmath}
\begin{dmath}[number={10l}]
\label{eq10l}
\thickmuskip=0mu
\medmuskip=0mu
\thinmuskip=0mu
= \frac{1}{\mathbb{E}\bigl[M^k(i)\bigr]\bigl(1-\mathbb{E}\bigl[M^k(i)\bigr]\bigr)}\sum_{m^k}\Bigl\{\Bigl(f(m^k\odot X)-\mathbb{E}\bigl[f(M^k\odot X)\bigr] \Bigr)\Bigl(m^k(i)-\mathbb{E}\bigl[M^k(i)\bigr] \Bigr)\Bigr\}P\Bigl[{M^k=m^k}\Bigr].
\end{dmath}
\end{dgroup*}

\noindent By the definition of covariance, we finally rewrite the summation as the expected values over $M^k$.
\begin{dmath}[number={11}]
\thickmuskip=0mu
\medmuskip=0mu
\thinmuskip=0mu
\tilde{F}_{i,k} \approx \frac{\mathbb{E}\bigl[f(M^k\odot X)\cdot M^k(i)\bigr]-\mathbb{E}\bigl[f(M^k\odot X)\bigr]\cdot \mathbb{E}\bigl[M^k(i)\bigr]}{\mathbb{E}\bigl[M^k(i)\bigr]\bigl(1-\mathbb{E}\bigl[M^k(i)\bigr]\bigr)}.\\
\end{dmath}

\subsection{Pseudocode}
The pseudocode describing our method is shown in Algorithm~\ref{alg:1}.
\begin{algorithm*}[h]
\newcommand{\multiline}[1]{%
  \begin{tabularx}{\dimexpr\linewidth-\ALG@thistlm}[t]{@{}X@{}}
    #1
  \end{tabularx}
}
\renewcommand{\algorithmicrequire}{\textbf{Inputs:}}
\renewcommand{\algorithmicensure}{\textbf{Outputs:}}
\caption{Pseudocode for computing attribution map $A$}
\begin{algorithmic}[1]
\Require{The number of masks $N$, number of layers $K$, detector function $\phi$, patch size $Z (= c \times c)$, image $X$ with a size of $H\times W$, explanation target detection ${\bm d_t}$, similarity calculation function $\mathcal{S}$, and all-ones matrix $J$.}
\Ensure{Attribution map $A$}
	\State $h \gets \lceil \frac{H}{c}\rceil, w \gets \lceil \frac{W}{c}\rceil, A\gets O$
	\For {$k=1, \ldots, K$}
		\State $p \gets \frac{k}{K+1}, A^k \gets O$
		\State $\text{sum\_score} \gets 0, \text{sum\_mask} \gets O, \text{sum\_score\_mask} \gets O$
		\For {$j=1, \ldots, N$}
		\State $M^k_j \gets$ \parbox[t]{\dimexpr\linewidth-\algorithmicindent}{%
		Generate a binary mask with a size of $h \times w$, where the elements are selected as 1 with probability $p$.\\ Subsequently, expand it to the size of $H \times W$ by the bilinear interpolation. %
		}
		\State $f\bigl({M^k_j}\odot X\bigr) \gets \max_{{\bm d_j} \in \phi(M^k_j\odot X)}
\mathcal{S}({\bm d_t}, {\bm d_j}) $
		\State $\text{sum\_score} \gets \text{sum\_score} + f\bigl({M^k_j}\odot X\bigr)$
		\State $\text{sum\_mask} \gets \text{sum\_mask} + M^k_j$
		\State $\text{sum\_score\_mask} \gets \text{sum\_score\_mask} + f\bigl({M^k_j}\odot X\bigr)M^k_j$
		\EndFor

	\State $\overline{f\bigl({M^k}\odot X\bigr)} \gets \text{sum\_score}/N$
	\State $\overline{M^k} \gets \text{sum\_mask}/N$
	\State $\overline{f\bigl({M^k}\odot X\bigr)M^k} \gets \text{sum\_score\_mask}/N$
	\State $A^k \gets \Bigl\{ \overline{f\bigl({M^k}\odot X\bigr)M^k}- \overline{f\bigl({M^k} \odot X\bigr)}\cdot \overline{M^k} \Bigr\} \oslash \Bigl\{ K\cdot Z\cdot \overline{M^k}\odot \bigl(J-\overline{M^k}\bigr)\Bigr\} $
	\State $A \gets A + A^k$
	\EndFor
	\State \Return $A$

\end{algorithmic}
\label{alg:1}
\end{algorithm*}

\twocolumn

\bibliography{aaai24}

\end{document}